\documentclass{elsarticle}

\makeatletter
\def\ps@pprintTitle{%
 \let\@oddhead\@empty
 \let\@evenhead\@empty
 \def\@oddfoot{}%
 \let\@evenfoot\@oddfoot}
\makeatother

\usepackage{times}
\usepackage{latexsym}
\usepackage{graphicx}
\usepackage{subcaption}

\usepackage[T1]{fontenc}

\usepackage[utf8]{inputenc}

\usepackage{microtype}

\usepackage{array}
\usepackage{pifont}
\usepackage{tabularx}
\usepackage{adjustbox}
\usepackage{multirow}
\usepackage{enumitem}
\usepackage{xspace}
\usepackage{tcolorbox}
\usepackage{booktabs,amsfonts,dcolumn}
\usepackage{hyperref}
\usepackage{url}
\usepackage{amsmath,amsthm,amsfonts,amssymb,bm,stmaryrd,bbm}
\usepackage[noorphans,vskip=0.75ex,leftmargin=2ex]{quoting}

\usepackage{xcolor}
\usepackage{cleveref}

\let\oldparagraph\paragraph
\renewcommand{\paragraph}[1]{\vspace{1em}\oldparagraph{#1}}

\title{Efficiency at Scale: Investigating the Performance of Diminutive Language Models in Clinical Tasks}

\begin{document}

\begin{frontmatter}

\author[inst1]{Niall Taylor*}
\author[inst1,inst2]{Upamanyu Ghose*}
\author[inst3,inst5]{Omid Rohanian} 
\author[inst4,inst5]{Mohammadmahdi Nouriborji}
\author[inst1]{Andrey Kormilitzin}
\author[inst3,inst6]{David A. Clifton}
\author[inst1]{Alejo Nevado-Holgado}

\affiliation[inst1]{organization={Department of Psychiatry, University of Oxford},
            city={Oxford},
            country={United Kingdom}}
\affiliation[inst2]{organization={Centre for Artificial Intelligence in Precision Medicines, University of Oxford and King Abdulaziz University},
            }
\affiliation[inst3]{organization={Department of Engineering Science, University of Oxford}, 
            city={Oxford},
            country={United Kingdom}}            
\affiliation[inst4]{organization={Sharif University of Technology}, 
            city={Tehran},
            country={Iran}}
\affiliation[inst5]{organization={NLPie Research}, 
            city={Oxford},
            country={United Kingdom}}
\affiliation[inst6]{organization={Oxford-Suzhou Centre for Advanced Research}, 
            city={Suzhou},
            country={China}}

\begin{abstract}
The entry of large language models (LLMs) into research and commercial spaces has led to a trend of ever-larger models, with initial promises of generalisability, followed by a widespread desire to downsize and create specialised models without the need for complete fine-tuning, using Parameter Efficient Fine-tuning (PEFT) methods. We present an investigation into the suitability of different PEFT methods to clinical decision-making tasks, across a range of model sizes, including extremely small models with as few as $25$ million parameters. 

Our analysis shows that the performance of most PEFT approaches varies significantly from one task to another, with the exception of LoRA, which maintains relatively high performance across all model sizes and tasks, typically approaching or matching full fine-tuned performance. The effectiveness of PEFT methods in the clinical domain is evident, particularly for specialised models which can operate on low-cost, in-house computing infrastructure. The advantages of these models, in terms of speed and reduced training costs, dramatically outweighs any performance gain from large foundation LLMs. Furthermore, we highlight how domain-specific pre-training interacts with PEFT methods and model size, and discuss how these factors interplay to provide the best efficiency-performance trade-off. Full code available at: tbd.
\end{abstract}
\end{frontmatter}

\section{Introduction} 
The Natural Language Processing (NLP) research space is now dominated by Large Language Models (LLMs), with a steady influx of different so-called foundation models from major AI companies every few months. The vast majority of recent LLMs are designed for \textit{generative} tasks and chat-style interactions, reliant on a mixture of autoregressive LM pre-training with follow-up reinforcement learning from human feedback (RLHF) to create the likes of ChatGPT \cite{openai_gpt-4_2023, touvron_llama_2023}.  However, the performance of these generative LLMs on classic NLP tasks such as sequence classification, relation extraction, named entity recognition, and embedding similarity search, especially in the clinical domain remains lacklustre \cite{moradi_gpt-3_2021, tunstall_efficient_2022, gutierrez_thinking_2022, sun_text_2023, tang_does_2023, rohanian_exploring_2023}. In many such cases, much smaller, BERT-style LLMs trained with masked language modelling (BERT, RoBERTa) continue to be competitive, or even surpass the performance of their larger counterparts \cite{chen_large_2023, rohanian_exploring_2023}. Moreover, achieving high performance with general domain LLMs on specialised clinical texts requires further adaptation through either extended pre-training on clinical data or fine-tuning for specific tasks.

\subsection{Scales of LLM}
Recent LLM research has predominantly focused on exceptionally large models from the more prolific AI companies, including ChatGPT from OpenAI \cite{openai_gpt-4_2023} and Llama \cite{touvron_llama_2023} from Meta. Although recent models from OpenAI are proprietary, it is widely recognised that the size of foundation models spans a broad range, from about $3$ to $175$ billion parameters, and with GPT-4 potentially more than one trillion parameters.  In contrast, there exist smaller, earlier-generation LLMs like RoBERTa-base, which contains approximately $125$ million parameters. The relative cost, simplicity, and reusability of these variously scaled models are crucial aspects to consider, and we aim to provide a holistic analysis of the interplay between different efficiency metrics and model size.

\subsection{Fine-tuning and PEFT}
Even smaller LLMs are relatively compute-intensive when compared to simpler machine learning alternatives, such as TF-IDF or Bag-of-Words paired with random forest classifiers. Moreover, adapting very large LLMs to new tasks can become unfeasible in low-resource settings where GPUs are scarce or non-existent. Common approaches to reduce model size include: knowledge distillation \cite{hinton_distilling_2015,sanh_distilbert_2020}, architecture compression \cite{sun_mobilebert_2020}, and pruning \cite{frantar_sparsegpt_2023}. These approaches generally aim to maintain a high level of performance in compressed models by harnessing the knowledge from the much larger \textit{teacher} LLMs. Whilst these approaches have had great success in producing smaller LLMs, adapting to new tasks still requires full fine-tuning of all model parameters to achieve optimal performance. This may necessitate a plethora of domain or task-specific LLMs, which cannot be used interchangeably due to catastrophic forgetting.\cite{luo_empirical_2023}. A more prevalent approach today is to adapt the fine-tuning approach itself.
Traditional approaches to adapting LLMs to downstream tasks involve introducing task specific neural network layers (often referred to as heads) to provide the extra flexibility required to complete a task, such as sequence classification. This training occurs in a supervised manner, involving updates to all model parameters, including task-specific ones (full fine-tuning). Full fine-tuning of smaller LLMs, such as BERT-base \cite{devlin_bert_2019} with merely $108$ million parameters has been feasible with modern GPUs, requiring only a single GPU with full precision. However, with the advent of models like Llama-2 \cite{touvron_llama_2023} with $65$ billion parameters, the practicality of fine-tuning these models on low-end hardware dwindles.

Several strategies exist to address this issue, one approach being the reduction of model size in terms of floating-point precision, bits, and the physical memory needed to store the weights through quantisation. This enables full fine-tuning of moderately sized models. \cite{dettmers_qlora_2023}. Pruning model parameters to reduce the \textit{redundant} weights for given downstream tasks has also been effective in certain cases \cite{frantar_sparsegpt_2023}. Another approach is to avoid full fine-tuning altogether, opting instead for zero-shot task adaption through prompting (prompt engineering), or by reducing the number of trainable parameters necessary for fine-tuning the LLM for its new task, a process known as Parameter Efficient Fine-tuning (PEFT). Notable PEFT methods include: Prompt tuning \cite{lester_power_2021}, Prefix tuning \cite{li_prefix-tuning_2021}, Low Rank Adaptation (LoRA) \cite{hu_lora_2021}, and Inhibit Activations ($IA^3$) \cite{liu_few-shot_2022}. These PEFT methods have become popular across various NLP tasks, and in this work, we will explore the utility of a select few for differently sized LLMs in the clinical domain.

\subsection{Clinical domain - LLM adaptation}
Unstructured clinical notes form a large portion of Electronic Health Records (EHRs) and can offer a substantial amount of clinically salient information given appropriate representation, such as that given by a LLM. Foundation LLMs are typically developed and trained for broad-stroke, general-purpose set of applications: trained on open, web-based text data and intended to be applied to \textit{similar} open, web-based text data. When taking foundation LLMs and applying to biomedical and clinical texts, performance often drops significantly \cite{alsentzer_publicly_2019, lehman_we_2023, moradi_gpt-3_2021, chen_large_2023, tunstall_efficient_2022, gutierrez_thinking_2022, sun_text_2023, tang_does_2023, yu_open_2023}. Achieving state-of-the-art (SoTA) performance in the clinical domain still involves training generic LLMs on biomedical or clinical domain data, and PEFT methods can provide efficient ways to adapt open LLMs to the clinical domain. The clinical domain is also inherently a compute-limited environment, with sensitive data which typically cannot be sent to third-party APIs. Thus, small, efficient LLMs that can perform specific tasks well and potentially run on edge devices are highly sought after \cite{rohanian_lightweight_2023, yu_open_2023}. 

\subsection{Related work}
Recent efforts have extensively explored the use of PEFT methods for large-scale models, aiming to align them with new domains or tasks \cite{dettmers_qlora_2023, ding_parameter-efficient_2023, hu_lora_2021}. However, despite the use of  quantisation and PEFT methods, high-end GPUs are still required and taking these models to production in any real-time setting becomes non-trivial in terms of cost and time.  
One group has recently investigated PEFT for clinical tasks with Llama models, and our work follows a very similar path \cite{gema_parameter-efficient_2023}. However, our emphasis is on the efficiency of these methods and how applicable they are to much smaller LLMs.

Our key contributions are:
\begin{itemize}
    \item Comparison of recent PEFT methods to clinical decision tasks
    \item The suitability of PEFT methods for small LLMS (Mobile and TinyBert architectures)
    \item The suitability of PEFT methods to knowledge distilled LLMs (DistilBERT)
    \item Exploring the interaction of pre-training domain, sample size and PEFT methods
\end{itemize}

\section{Methods}

\subsection{Model architectures}
We evaluate the performance of PEFT across various transformer-based LLM architectures of differing sizes, including: TinyBERT \cite{jiao_tinybert_2020}, MobileBERT \cite{sun_mobilebert_2020}, DistilBERT \cite{sanh_distilbert_2020} , standard BERT \cite{devlin_bert_2019}, and Llama-2-7b \cite{touvron_llama_2023}. A table of relevant architecture details is provided in Table \ref{tab:model-details}. 

\begin{table}
    \centering
    \footnotesize
    \begin{tabular}{cccc}
    \toprule
        \textbf{Model architecture} & \textbf{\# Params (mil)}  & \textbf{GPU (VRAM GB}) &\textbf{ FLOPs} \\
        \midrule
         Tiny-BERT & 13.87 & 0.052 & $3.66 \times 10^{7}$\\
         Mobile-BERT & 24.58 & 0.092  & $1.62 \times 10^{8}$ \\
         Distil-BERT & 65.78  & 0.245  & $3.41 \times 10^{8}$ \\
         BERT & 108.31 & 0.403  & $6.81 \times 10^{8}$\\
         Llama2-7b &  6607.34 & 24.6   & $5.18 \times 10^{10}$\\
         Llama2-7b (bfloat16) & 6607.34 & 12.37  & $5.18 \times 10^{10}$\\
         \bottomrule

    \end{tabular}
    \caption{Model architectures and their associated number of parameters, Video Random Access Memory (VRAM), and Floating Point Operations (FLOPs). FLOPs were based on a random sample of 10 tokens.}
    \label{tab:model-details}
\end{table}

\subsection{Domain pre-training}
In addition to exploring various transformer-based LLM architectures of different sizes, we examine three domain variants for each: 

\begin{itemize}
\item \textbf{General:} Original, unadapted models.
\item \textbf{Biomedical:} Models pre-trained or distilled with biomedical literature \cite{rohanian_effectiveness_2023}
\item \textbf{Clinical:} Models pre-trained with clinical EHR data \cite{rohanian_lightweight_2023}
\end{itemize}

This framework allows us to investigate the interplay between domain pre-training, model size, and the chosen PEFT methods.

\subsection{Downstream fine-tuning}

We opt to compare performance using a traditional fine-tuning setup, whereby each LLM is adapted with a task-specific head to perform the respective downstream task. For each task, we will utilise additional linear layers on top of the base LLM, with a task-specific loss that is used to update all model parameters (the base LLM and the additional task head). This approach remains the most suitable across all model architectures and aligns with previous research \cite{taylor_clinical_2023, rohanian_lightweight_2023}.

\subsection{PEFT}

Parameter Efficient Fine-tuning (PEFT) methods are numerous, but they typically fall into two categories: introducing new trainable parameters or selectively freezing existing ones. For our experiments, we focus on the following methods. In addition to the trainable parameters specific to each method described below, the task-specific parameters in the classification head are also trained.

\paragraph{Low-Rank Adaptation of Large Language Models}
Low-Rank Adaptation of LLMs or LoRA \cite{hu_lora_2021} is a reparameterisation technique that works by injecting two trainable matrices ($A$ and $B$) that act as an approximation of a singular value decomposition (SVD) of the weight update $\Delta W$ for any weight matrix $W \in \mathbb{R}^{d \times k}$ in the LLM. The approximation works as $\Delta W = BA$, where $B \in \mathbb{R}^{d \times r}$, $A \in \mathbb{R}^{r \times k}$ and $r \ll min(d, k) $ is the rank of the LoRA matrices, which is a tunable parameter. The new forward pass is updated to $h = (W + \Delta W)x = (W + AB)x = Wx + ABx$. While it is possible to introduce the LoRA matrices in any layer of the LLM, it is common practice to introduce them as weight update approximations for the key, query and value matrices. The underlying assumption is that the weight updates in LLMs intrinsically have a lower rank than their dimensions, and thus can be well approximated by their SVD. Additionally, once fully trained, the LoRA matrices can be integrated into the model as $W_{updated} = W_0 + BA$, thereby introducing no inference latency. With LoRA the original weight matrices of the LLM remain frozen during the fine-tuning phase. 

\paragraph{$IA^3$}
Infused Adapter by Inhibiting and Amplifying Inner Activation ($IA^3$) shares similarities with other adapter methods that introduce new parameters to scale activations using learned vectors \cite{liu_few-shot_2022}. While these learnable vectors can be applied to any set of activations, applying them to the keys and values in the relevant attention mechanism and the intermediate activation of the position-wise feed-forward networks was found to be both efficient and sufficient. For a transformer based architecture, we have a key $K\in\mathbb{R}^{d_k}$ and value $V\in\mathbb{R}^{d_v}$, and the hidden dimensions of the position-wise feed-forward network is $d_{ff}$. $IA^3$ introduces learnable vectors $l_k\in\mathbb{R}^{d_k}$, $l_v\in\mathbb{R}^{d_v}$ and $l_{ff}\in\mathbb{R}^{d_{ff}}$ and modifies the attention and feed-forward calculation as follows:

\begin{equation}
\textit{softmax}\left( \frac{Q(l_k \odot K)}{\sqrt{d_k}} \right)(l_v \odot V)
\end{equation}
\begin{equation}
(l_{ff} \odot \gamma(W_1 x)) W_2    
\end{equation}
where $\odot$ represents the element-wise product, and $\gamma$, $W_1$ and $W_2$ are the activation function and weight matrices of the feed-forward network. Similar to LoRA, the learnable vectors can be merged into the model as $l \odot W$ because any operation $l \odot Wx$ is equivalent to $(l \odot W)x$. Hence, this method does not introduce any inference latency either. Once again, with $IA^3$ the original weight matrices of the LLM remain frozen during fine-tuning.

Based on previous works and some preliminary experiments, we opt to focus on LoRA and $IA^3$ for our main experiments, which generally demonstrate significantly better performance compared to alternative PEFT methods. Moreover, aligning prefix tuning and prompt learning with NER tasks is not straightforward and we believed it offered limited value to adapt these methods for NER specifically (for a comparison of other PEFT methods, see previous work\cite{gema_parameter-efficient_2023}). 

\subsection{Few-Shot  training}
A prevalent challenge in real-world scenarios  is the scarcity of training samples, especially in the clinical domain where certain diseases are inherently rare and generating gold-standard annotations demands clinical expertise and considerable time, both of which are limited resources. Therefore, the ability to train a viable model with few training samples is another angle of efficiency we explore. This is achieved by supplying only a limited number of training samples per class to a specific model. We carry out a series of experiments with an escalating number of samples per class to determine the effect of different model sizes and PEFT methods.

\subsection{Datasets and Tasks}
We utilise a number of commonly used clinical datasets for downstream evaluation, focusing on the following tasks: named entity recognition (NER), sequence classification and relation extraction (RE), in line with earlier clinical NLP research \cite{meehan_clinical_2022, lee_biobert_2020}. 

\subsubsection{Sequence classification tasks}
\paragraph{MIMIC-III ICD-9 Triage}
A common task with the MIMIC-III dataset \cite{johnson_mimic-iii_2016} involves classifying patient records according to their medical diagnoses, which are coded using a system known as ICD-9. We utilise a simplified version of this task, where the top $20$ most commonly occurring ICD-9 codes are categorised into seven \textit{triage} groups: [\textit{Cardiology, Obstetrics, Respiratory, Neurology, Oncology, AcuteMedicine, Gastroenterology}]. This grouping was developed in collaboration with clinicians. For further information, please refer to the original paper \cite{taylor_clinical_2023}.

 \paragraph{MIMIC-III - Clinical Outcomes}

Two clinical outcome tasks associated with the MIMIC-III dataset \cite{johnson_mimic-iii_2016} are Mortality Prediction (MP) and Length of Stay (LoS) prediction \cite{van_aken_clinical_2021}. MP involves analysing discharge summaries from the ICU to assess a patient's mortality risk, constituting a binary classification problem. The LoS task also uses ICU discharge summaries to forecast the duration of a patient's hospital stay, with durations binned into four classes: under $3$ days, $3$ to $7$ days, $1$ week to $2$ weeks, and more than $2$ weeks.

\paragraph{I2B2 2010 Relation Extraction}
We used several curated datasets from the I2B2 series, including the 2010 medical relation extraction dataset \cite{uzuner_2010_2011} which aims to classify text based on the apparent medical relationship being described, with the following derived labels:

\begin{enumerate}
    \item Treatment improves medical problem (TrIP)
    \item Treatment worsens medical problem (TrWP)
    \item Treatment causes medical problem (TrCP)
    \item Treatment is administered for medical problem (TrAP)
    \item Treatment is not administered because of medical problem (TrNAP)
    \item Test reveals medical problem (TeRP)
    \item Test conducted to investigate medical problem (TeCP)
    \item Medical problem indicates medical problem (PIP)
    \item No Relations
\end{enumerate}

We follow the same pre-processing procedure outlined in previous works \cite{rohanian_lightweight_2023}.

\subsubsection{Named Entity Recognition}
\paragraph{I2B2 - 2010 and 2012}

These two NER tasks involve classifying text spans related to temporal relations \cite{uzuner_2010_2011, sun_evaluating_2013} within discharge summaries, as delineated by expert annotations. The classification is based on four primary categories: clinical concepts, clinical departments, evidentials, and occurences. These categories are further broken down into more specific entities: \textit{medical problem (PR)}, \textit{medical treatment (TR), medical test (TE)}, \textit{clinical department (CD), evidential (EV)}, \textit{occurence (OC)}, and \textit{none (NO)}. 

\paragraph{I2B2 - 2014}
A deidentification task, whereby spans of text within clinical notes are classified using different protected health information (PHI) such as name, address, and postcode \cite{stubbs_automated_2015}.

\begin{table*}
    \centering
    \footnotesize
    \begin{tabular}{ccccc}
    \toprule
        \textbf{Dataset} & \textbf{Task Type} & \textbf{\# labels} & \textbf{\# train samples} & \textbf{\# eval samples} \\
        \midrule
        MIMIC-III MP & Seq. CLS  & 2 & 33,954 & 9,822 \\
        MIMIC-III LoS & Seq. CLS & 3 & 30,421 & 8,797 \\
        MIMIC-III ICD-9 Triage & Seq. CLS & 7 & 9,559 & 3,172 \\
         I2B2 2010 RE & Seq. CLS & 9 & 22,256 & 43,000\\
         I2B2 2010   & NER  & 7 & 6726 & 27,626 \\
         I2B2 2012  & NER & 13 & 6797  &  5,664\\
         I2B2 2014  & NER & 42 & 45974 & 32,586\\
         \bottomrule
    \end{tabular}
    \caption{Dataset details.}
    \label{tab:dataset-details}
\end{table*}

For further dataset and task details, see \ref{appendix:dataset-details}.
\section{Results}
\subsection{Model size vs PEFT}
\label{results:model-size-peft}
The number of trainable parameters is an important factor in determining the efficiency in model performance and has a strong correlation with cost and time of training. We detail the performance metrics for various PEFT methods applied to each model type across different clinical tasks. In Table \ref{tab:cls-ner-results}, we present the results for sequence classification and NER across different PEFT methods and model sizes.

\begin{table}[htp]

\centering
\begin{subtable}{\textwidth}
  \centering
  \footnotesize
  \begin{tabular}{llcccc}
    \toprule  
        \multicolumn{1}{c}{\bf Model name} & \bf PEFT & \multicolumn{1}{c}{\bf ICD9-Triage} & \multicolumn{1}{c}{\bf i2b2-2010-RE} & \multicolumn{1}{c}{\bf MIMIC-LoS} & \multicolumn{1}{c}{\bf Mimic-MP} \\
        \midrule
        BioBERT & Full & \underline{0.864} (0.002) & \underline{0.935} (0.004) & \underline{0.709} (0.002) & \underline{0.819} (0.020) \\
        & IA3 & 0.703 (0.19) & 0.896 (0.004) & 0.634 (0.001) & 0.769 (0.005) \\
        & \textbf{LORA} & \textbf{0.827} (0.002) & \textbf{0.925} (0.001) &\textbf{ 0.697} (0.002) & \textbf{0.828} (0.002) \\   
        \midrule
        BioDistilBERT & Full & 0.862 (0.010) & 0.927 (0.003) & \underline{0.706} (0.003) & 0.825 (0.006) \\  
        & IA3 & 0.792 (0.008) & 0.906 (0.002) & 0.677 (0) & 0.797 (0.001) \\
        & \textbf{LORA} & \textbf{0.855} (0.005) & \textbf{\underline{0.928}} (0.003) & \textbf{0.702} (0.001) & \textbf{0.825} (0.001) \\  
        \midrule
        BioMobileBERT & Full & \underline{0.851} (0.004) & \underline{0.932} (0.003) & \underline{0.704} (0.004) & \underline{0.819} (0.011) \\
        & IA3 & 0.744 (0.012) & 0.897 (0.003) & 0.639 (0.001) & 0.774 (0.002) \\  
        & \textbf{LORA} & \textbf{0.808} (0.004) & \textbf{0.918} (0.002) & \textbf{0.671} (0.004) & \textbf{0.798} (0.002) \\
        \midrule
        TinyBioBERT & Full & \underline{0.727} (0.012) & \underline{0.910} (0.005) & \underline{0.684} (0.001) & \underline{0.802} (0.001) \\   
        & IA3 & 0.390 (0.035) & 0.852 (0.002) & 0.588 (0.003) & 0.607 (0.003) \\
        & \textbf{LORA} & \textbf{0.599} (0.008) & \textbf{0.895} (0.003) & \textbf{0.649} (0.006) & \textbf{0.764} (0.003) \\ 
        \bottomrule
  \end{tabular}
  \caption{Sequence classification task results}
  \label{tab:cls_results}
\end{subtable}

\vspace{0.5cm} 

\begin{subtable}{\textwidth}
  \centering
  \footnotesize
 \begin{tabular}{llrrr}
   \toprule
        \multicolumn{1}{c}{\bf Model name} & \bf PEFT & \multicolumn{1}{c}{\bf i2b2-2010-NER} & \multicolumn{1}{c}{\bf i2b2-2012-NER} & \multicolumn{1}{c}{\bf i2b2-2014-NER} \\
        \midrule
        BioBERT & Full & \underline{0.819} (0.003) & \underline{0.824} (0.001) & \underline{0.967} (0.001) \\
                & IA3 & 0.473 (0.002) & 0.485 (0.006) & 0.850 (0.001) \\  
                & LORA & \textbf{0.696} (0.003) & \textbf{0.753} (0.001) & \textbf{0.935} (0) \\
        \midrule
        BioDistilBERT & Full & \underline{0.803} (0.003) & \underline{0.795} (0.006) & \underline{0.967} (0.001) \\   
                    & IA3 & 0.498 (0.003) & 0.503 (0.001) & 0.883 (0) \\
                    & LORA & \textbf{0.718} (0.008) & \textbf{0.729} (0.006) & \textbf{0.940} (0.001) \\
        \midrule 
        BioMobileBERT & Full & \underline{0.796} (0.003) & \underline{0.772} (0.006) & \underline{0.966} (0) \\  
                    & IA3 & 0.515 (0.003) & 0.515 (0.003) & 0.908 (0) \\ 
                    & LORA & \textbf{0.638} (0.010) & \textbf{0.650} (0.004) & \textbf{0.941} (0.001) \\
        \midrule
        TinyBioBERT & Full & \underline{0.655} (0.004) & \underline{0.705} (0.008) & \underline{0.906} (0.003) \\  
                    & IA3 & 0.328 (0.009) & 0.381 (0.003) & 0.715 (0.002) \\ 
                    & LORA & \textbf{0.438} (0.007) &\textbf{ 0.561} (0.009) & \textbf{0.8051} (0.013) \\
        \bottomrule
 \end{tabular}
 \caption{NER task results}
 \label{tab:ner_results}
\end{subtable}
\caption{PEFT results for all downstream tasks using biomedical models, with values representing the median from 3 distinct training runs under varied random seeds for PyTorch weight initialisations. Standard Deviation ($SD$) is provided in brackets. Micro-averaged F1 scores are reported for the i2b2-2010-RE and all NER tasks. Macro-averaged Receiver Operating Characteristic area under the curve ($ROC AUC$) is used for MIMIC-LoS and MP tasks, while macro-averaged F1 scores are reported for the ICD-9 triage task. \textbf{Bold} results indicate best PEFT performance, and values underlined are top performance across all fine-tuning methods.} 
\label{tab:cls-ner-results}
\end{table}

The results demonstrate that LoRA consistently outperforms other PEFT methods across all models and tasks, often approaching the performance of full fine-tuning.

\begin{figure*}
    \centering
    \includegraphics[scale=0.5, trim={0cm 0cm 0 0}]{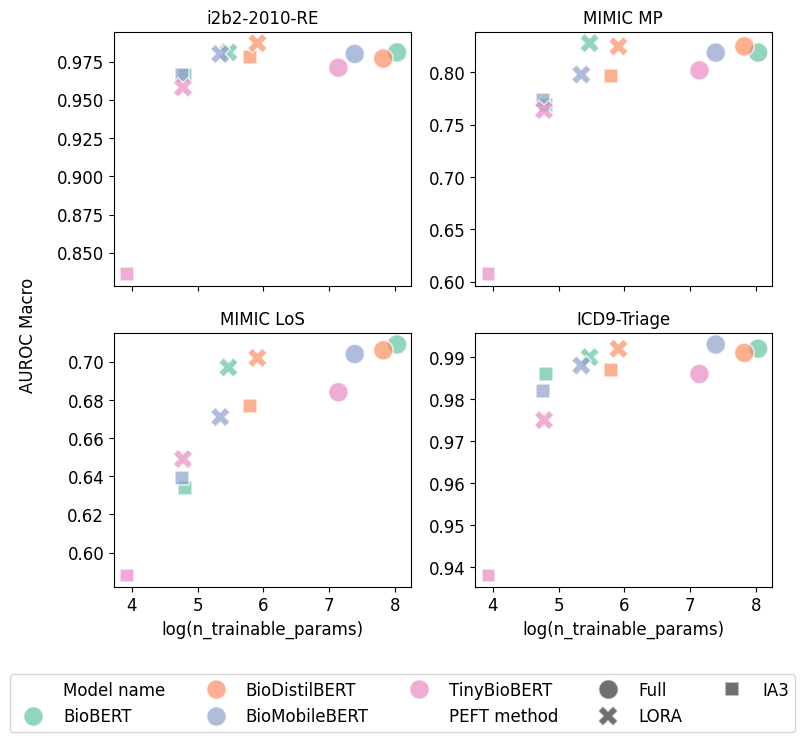}
    \caption{Sequence classification performance across the different LLM model sizes and the associated number of trainable parameters.}
    \label{fig:trainable-params-vs-performance}
\end{figure*}

We also present a comparison of the number of trainable parameters as a function of the different PEFT methods in Fig \ref{fig:trainable-params-vs-performance}. There is a clear correlation between the number of trainable parameters and performance, and LoRA appears to provide larger models an advantage over fully fine-tuned smaller models.

\subsection{Differential effect of LoRA rank according to model size}


Given the superior performance of LoRA over other PEFT methods, as evidenced in Figure \ref{fig:trainable-params-vs-performance}, we aimed to methodically evaluate the impact of the LoRA rank hyperparameter across models of varying sizes. For this purpose, we employed the Optuna package \cite{akiba_optuna_2019} to conduct 20 trials of hyperparameter optimisation, holding the LoRA rank constant at $r \in {8, 16, 32, 64, 128}$. The hyperparameters adjusted during tuning included LoRA dropout ($d \in {0.1, 0.3, 0.5}$), LoRA alpha ($\alpha \in {0.3, 0.5, 1.0}$), and learning rate ($lr \in [10^{-5}, 10^{-3}]$). The Llama model was excluded from this experiment due to its significantly larger size compared to BERT-based models, which would have imposed an excessive computational load for hyperparameter tuning. Following the hyperparameter search, we selected the optimal performing model for each $r$ value to analyse its effect on models with differing parameter counts (Appendix \ref{appendix:LoRA-rank-vs-performance}).

Increasing the rank $r$ in TinyBioBERT led to improved performance up to $r=64$, after which a slight decline was observed at $r=128$. A similar pattern was noted in BioDistilBERT, with the turning point at $r=32$. The impact of rank on BioMobileBERT was more variable, with a noticeable performance dip only at $r=64$. This variability might be attributed to the distinct architecture of BioMobileBERT compared to other BERT-based models \cite{sun_mobilebert_2020}. For BioBERT, the larger model in the BERT family, there was a modest improvement at $r=16$, but performance tended to decrease at higher ranks. Conversely, for the RoBERTa model, performance enhancements were seen at ranks $r=32$ and $r=128$, yet no clear pattern between rank and performance emerged. Despite these fluctuations, the overall impact on model performance was relatively minor, with the greatest increase in AUROC being $0.0125$ and the largest decrease being $0.0078$. Hence, even for models with varying number
of parameters, the default LoRA rank of $8$ is a good trade-off between computational time taken to tune the models and performance. However, if the task at hand would practically benefit from a small increase in the performance metric, tuning the LoRA parameters may be beneficial. 

\subsection{General vs biomedical vs clinical domain pre-training}
Another aspect of efficiency with regards to LLM downstream adaptation is the domain in which the model was pre-trained. We have conducted direct comparisons between models pre-trained in general, biomedical, and clinical domains across our various model architectures. For the sake of brevity, we focus solely on the i2b2-2010 relation extraction task. The performance differences are greatest in the smaller models, with clinically pre-trained models generally performing best with a 1-4 percent improvement based on model size. For results across all tasks  and their dependence on domain pre-training, please see Appendix \ref{appendix:all-domain-results}.

\begin{figure}[htp]
    \centering
    \includegraphics[scale=0.5]{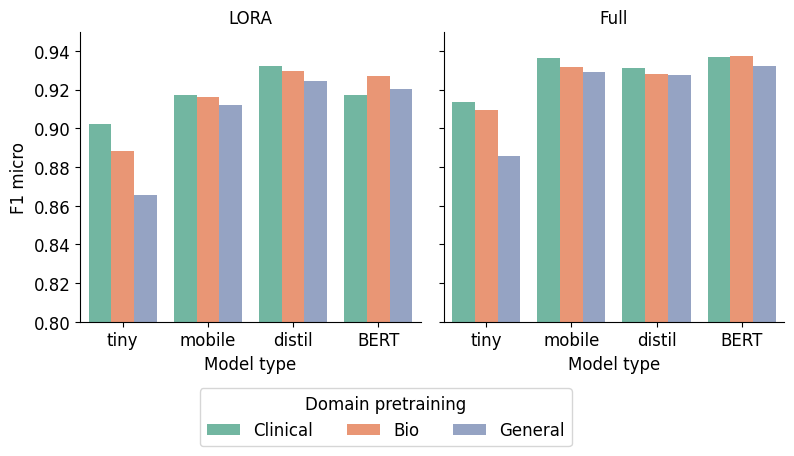}
    \caption{Comparison of F1 micro scores on the I2B2 2010 relation extraction task dependent on whether the model received biomedical, clinical, or general domain pre-training.}
    \label{fig:domain-vs-peft}
\end{figure}

\subsection{Budget}
The primary advantage of employing PEFT methods lies in their ability to reduce training times, lower GPU memory demands, minimise storage requirements, and enhance model reusability (all of which lower financial burden). In our study, we examined the trade-offs among these aspects for various model architectures, focusing on the most effective PEFT method identified in our experiments, namely, LoRA. For each defined budget, we used MIMIC mortality prediction as the benchmark task and macro-averaged AUROC as the metric of evaluation. In addition to training the LoRA versions of each model, we also conducted full fine-tuning on each model to determine whether any budget level could achieve efficiency improvements comparable to those provided by PEFT approaches. The only exception was the Llama model, which was exclusively trained with LoRA due to computational constraints.

\subsubsection{Time}


A key measure of efficiency is the training time and the speed at which different models converge within a constrained period, particularly a relatively short one. We set an initial time limit of $2,000$ seconds ($33$ minutes) for all models. To evaluate the performance of the models that seemed to show an increasing trend in performance after the budget of $2,000$ seconds (Figure \ref{fig:time-fewshot-budget}), we raised the budget to $6,000$ seconds ($100$ minutes). An exception was made for the Llama model, which remained under-trained even after $6,000$ seconds, necessitating an extension of the training period to approximately $21,500$ seconds ($6$ hours) to attain optimal performance. 

We observed that the fully fine-tuned version of the models, regardless of size, was quicker to converge than the LoRA versions, followed by eventually overfitting. The LoRA versions of the models eventually converged to the performance (or close to the performance) of the fully fine-tuned models. This observation suggests that fully fine-tuning a model on a small time budget could theoretically obtain an efficiency gain similar to the PEFT methods.  However, from a practical standpoint, the LoRA version of all models converged to similar performance within $\sim$1 hour of training (Figure \ref{fig:time-fewshot-budget}) while being more memory efficient. A more detailed analysis of the difference in efficiency between the methods is discussed in section \ref{results:memory_and_cost} It is also important to acknowledge that larger models, such as Llama, deliver superior performance but incur significantly higher time and memory costs. 

\subsubsection{Few-shot Training}


Another focus for efficient training involves restricting the number of training samples, reflecting real-world situations with especially rare outcomes or cases where producing labels is challenging. We explored sample budgets that ranged from $8$ to $4096$ samples, increasing incrementally by a factor of 2.

\begin{figure}[p]
        \centering
        \begin{subfigure}{.9\textwidth}
            \centering
            \caption{Time budget sensitivity}
            \includegraphics[scale=0.5,
                            trim={0 0 0 7},clip]{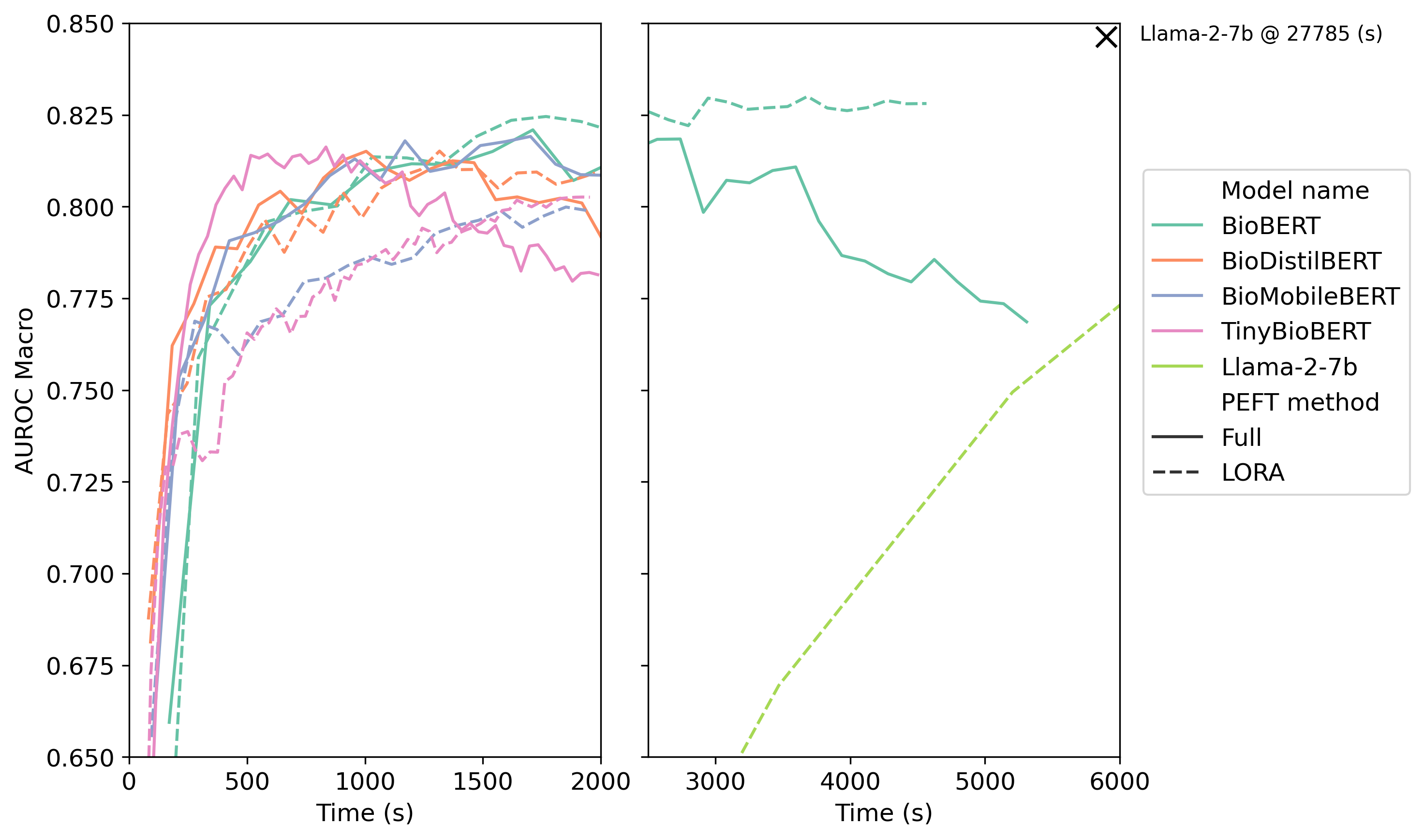}
        \end{subfigure}
        \hfill
        \begin{subfigure}{.9\textwidth}
            \centering
            \caption{Few-shot sensitivity}
            \includegraphics[scale=0.5]{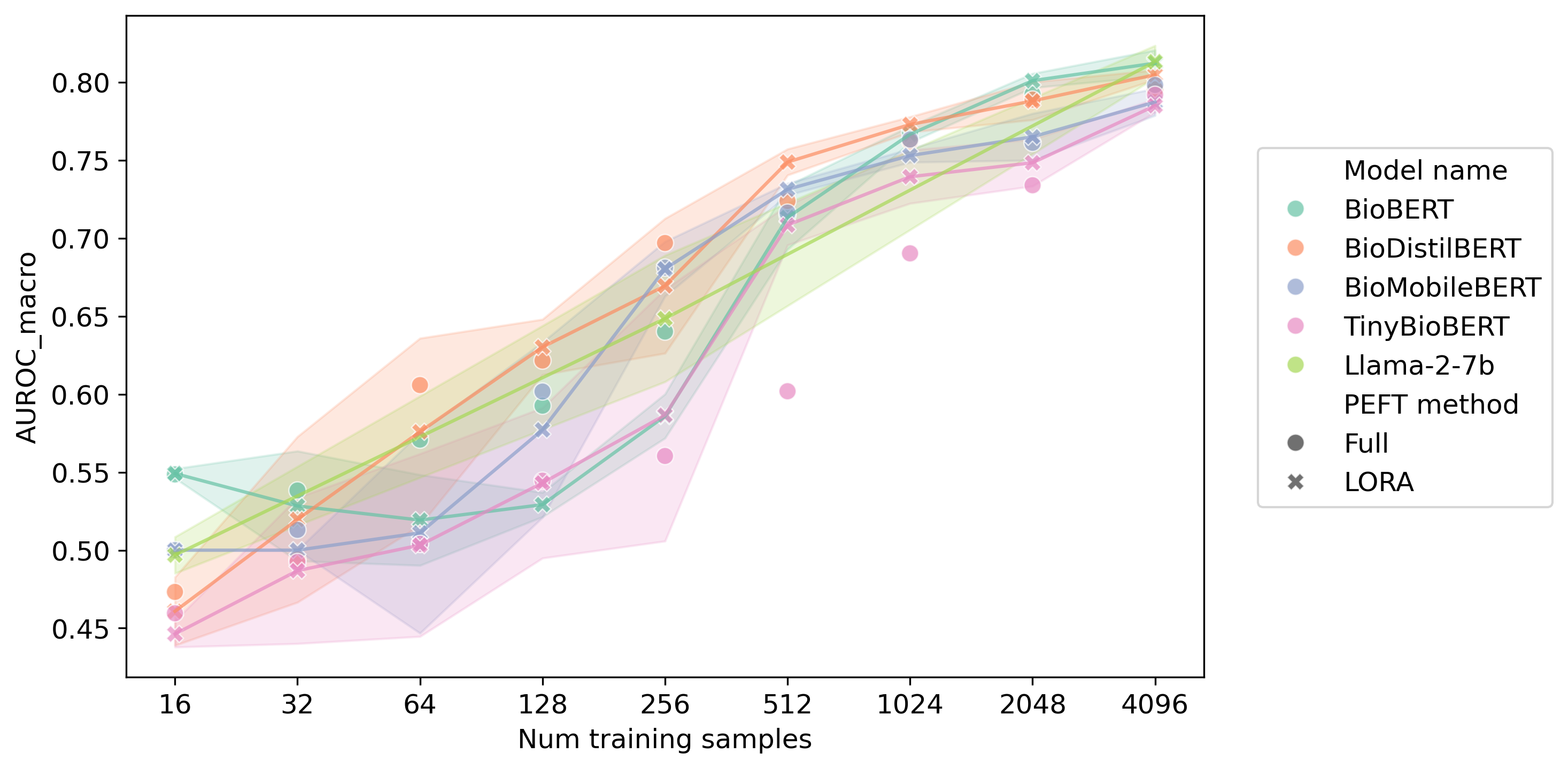}
        \end{subfigure} 
        \caption{Effect of training time (\textbf{a}) and few-shot sampling (\textbf{b}) on models of varying sizes, trained using full fine-tuning as well as LoRA. The connected points reflect the LoRA results to highlight the trend. The task used for this experiment was MIMIC mortality prediction.}
        \label{fig:time-fewshot-budget}
\end{figure}

As expected, we observed a direct relationship between  sample budget and model performance, regardless of the model type and training method used. While we noticed the fully fine-tuned models generally performing better than their LoRA counterparts for smaller sample budgets, the difference became negligible for higher budget values (Figure \ref{fig:time-fewshot-budget}). The fully fine-tuned models on a budget of $4096$ samples under performed when compared against the LoRA versions on all samples. Hence, for sample budget to be considered as an effective method for efficiency gain, we would need more than $4096$ samples.

\subsubsection{Holistic efficiency}

In an attempt to establish a unified metric of efficiency, we took the average of the following normalised metrics: time taken to reach peak performance $T$, number of trainable parameters $P$ and total model parameters $S$:

\begin{equation}
\text{Efficiency} = \frac{{T} + {P} + {S}}{3}  
\end{equation}

\begin{figure}[htp]
    \centering
    \includegraphics[scale=0.5]{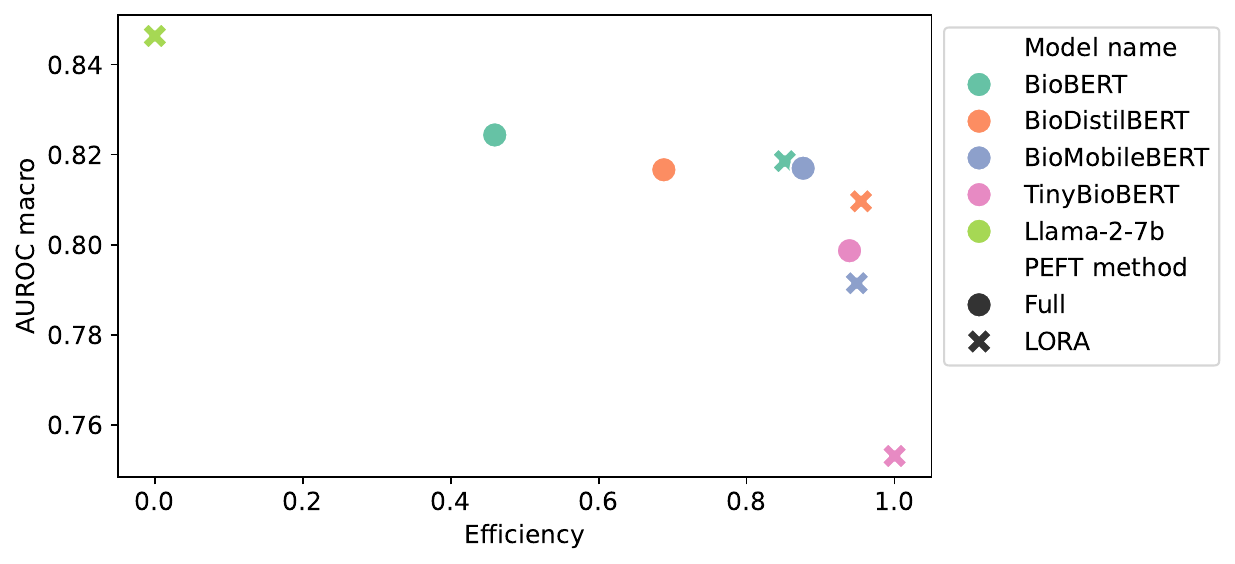}
    \caption{Comparison of efficiency against performance on the validation set between models of different size. }
    \label{fig:efficiency-plot}
\end{figure}

For ease of interpretability, we scaled the final efficiency value to range between 0 and 1, where 0 represents the least efficient model and 1 represents the most efficient. We show the relationship between efficiency and performance in Figure \ref{fig:efficiency-plot} \footnote{we note that there is a change in performance gap on the held-out test set between \textit{LoRA} and \textit{Full} compared to the validation set reported elsewhere}.

The holistic efficiency shows a general negative correlation between efficiency and performance, however the gap in performance is relatively minor compared to the difference in efficiency between models.

\subsubsection{Memory and cost}
\label{results:memory_and_cost}


\begin{table}[htp]
\centering
\footnotesize
\setlength{\tabcolsep}{5pt}

\begin{tabular}{@{}llrrr@{}}
\toprule
\multicolumn{1}{c}{\bf Model name} & \bf PEFT Method & \multicolumn{1}{c}{\bf Train time (hr)} & \multicolumn{1}{c}{\bf Inference time (hr)} & \multicolumn{1}{c}{\bf Total cost (GBP)} \\ 
\midrule
Llama-2-7b & LORA & 51.07 & 4.06 & 112.22 \\
\midrule  
BioBERT & Full & 2.51 & 0.22 & 5.56 \\
BioBERT & LORA & 2.16 & 0.22 & 4.84 \\  
\midrule
BioMobileBERT & Full & 1.57 & 0.14 & 3.48 \\ 
BioMobileBERT & LORA & 1.35 & 0.14 & 3.03 \\
\midrule
BioDistilBERT & Full & 1.35 & 0.12 & 2.99 \\  
BioDistilBERT & LORA & 1.21 & 0.13 & 2.73 \\ 
\midrule
TinyBioBERT & Full & 0.53 & 0.06 & 1.20 \\   
TinyBioBERT & LORA & 0.46 & 0.06 & 1.06 \\
\bottomrule
\end{tabular}  

\caption{Costs for training each model on a task with approximately 30,000 training samples for 10 epochs, followed by running it in inference mode for 100,000 samples. The costs were estimated using AWS EC2 rates. The instances used for estimating training and inference costs were g5.16xlarge and g4dn.16xlarge, respectively.}
\label{tab:aws_cost}
\end{table}

The GPU and storage requirements for training differ massively between model types, and fine-tuning method. Whilst performance has generally increased with model size, there is a trade-off between performance and compute required, as well as speed of training and inference. We provided the model size and memory requirements in Table \ref{tab:model-details} and we extend this analysis by calculating the estimated costs of training and storage of the differently sized models in Table \ref{tab:aws_cost}.  As observed in previous results, larger models like Llama-2-7b achieve higher performance on most tasks but at $20$ and $94$ times the monetary value of models like BioBERT and TinyBioBERT, respectively. If the objective is to fine-tune a model for multiple tasks, BioBERT and similar models can be a good trade-off between monetary cost and performance.

\section{Discussion}

\subsection{PEFT with small LLMs}

We have explored the use of different-sized LLMs for various clinical downstream tasks, assessing both traditional fine-tuning and different PEFT methods. From the methods we studied ($IA^3$ and \textit{LoRA}), we found LoRA to be superior across all tasks, leading us to select it as the preferred PEFT method for all subsequent analysis. Whilst full fine-tuning generally outperforms LoRA, in certain models and tasks the performance is at least matched or even surpassed and that LoRA works well for all model sizes. This finding highlights the potential in utilising PEFT methods with very small LLMs. The relative performance gap between full fine-tuning and LoRA appears to increase with the smaller models, which was only partially mitigated by increasing the LoRA rank.

\subsection{Comparison of LLM size}
The performance of various model sizes was evaluated on a specific task within a fixed time frame, including the $7$ billion parameter Llama-2 model. This comparison revealed significant differences in the learning capabilities of models of varying sizes. Numerous smaller LLMs completed $5$ epochs of training well before the Llama-2 Llama-2 model achieved comparable performance levels. Nevertheless, when given sufficient time, Llama-2 did reach the highest evaluation performance by a few percentage points in the target task. Llama-2 model is approximately $500$ times the size of the TinyBERT models, indicating that the computational demand, even with the implementation of LoRA for Llama-2, is significantly higher. The duration required for the Llama-2 model to achieve comparable performance on downstream tasks, using the same GPU, was considerable. It took roughly ten times longer to match the performance of smaller LLMs and exceeded six hours of training to attain its peak performance.

\subsection{Holistic efficiency}
According to our composite efficiency metric, the medium sized LLMs are substantially more computationally efficient compared to the largest model for the given task, whilst only exhibiting a minor drop in performance. It is difficult to derive a true representation of holistic efficiency as this would likely require taking cost and time of pre-training, and other facets not known, but we believe this provides a reasonable overview of the interplay between model size and fine-tuning methods. Further profiling would be needed to quantify exact runtime improvements.

\subsection{Domain pre-training}
The pre-training of LLMs proved quite important in the performance on the various clinical domain tasks, with biomedical and clinical LLMs generally outperforming their general counterparts. We do note that the \textit{clinical} LLMs, such as ClinicalBioBERT have been trained on MIMIC-III notes themselves and this does give them an unfair advantage. However, the potential for data leakage in the Llama-2 model is difficult to ascertain. In line with previous works \cite{lehman_we_2023}, it could be argued that developing specialised clinical LLMs through pre-training on relevant clinical language remains optimal for subsequent downstream task adaptation.

\subsection{Limitations and future work}
The selection of PEFT methods investigated in this study reflected the state of the field at the time; however, we acknowledge that this is an evolving research area, and we cannot be certain that other methods would not have outperformed those presented here. Indeed, since conducting these experiments, the PEFT library\cite{peft} has introduced several new methods worth exploring.

When comparing various model sizes, we chose to limit training to a single GPU. This approach might disadvantage larger models, particularly the Llama-2 model, which was forced to employ reduction in bit-precision to allow any training. Furthermore , this constraint hindered our ability to thoroughly investigate Llama-2 across all tasks and conduct any hyperparameter optimisation.  Future work could seek to explore this further, although the resources required are extensive and arguably  yield diminishing returns. 
\subsection{Conclusion}
Overall, we believe this work highlights the power of PEFT methods for small LLMs and demonstrates how domain pre-training can be leveraged to create efficient clinical models. While the capabilities of much larger LLMs are evident, they come with significantly higher time and financial demands.

\section*{Funding}
 NT was supported by the EPSRC Center for Doctoral Training in Health Data Science (EP/S02428X/1). UG was supported by Alzheimer's Research UK, and the Centre for Artificial Intelligence in Precision Medicines (University of Oxford and King Abdulaziz University). DAC was supported by the Pandemic Sciences Institute at the University of Oxford; the National Institute for Health Research (NIHR) Oxford Biomedical Research Centre (BRC); an NIHR Research Professorship; a Royal Academy of Engineering Research Chair; and the InnoHK Hong Kong Centre for Centre for Cerebro-cardiovascular Engineering (COCHE).

\bibliographystyle{unsrtnat}
\bibliography{other_references.bib, all_references.bib}

\begin{thebibliography}{39}
\providecommand{\natexlab}[1]{#1}
\providecommand{\url}[1]{\texttt{#1}}
\expandafter\ifx\csname urlstyle\endcsname\relax
  \providecommand{\doi}[1]{doi: #1}\else
  \providecommand{\doi}{doi: \begingroup \urlstyle{rm}\Url}\fi

\bibitem[OpenAI(2023)]{openai_gpt-4_2023}
OpenAI.
\newblock {GPT}-4 {Technical} {Report}, March 2023.
\newblock URL \url{http://arxiv.org/abs/2303.08774}.
\newblock arXiv:2303.08774 [cs].

\bibitem[Touvron et~al.(2023)Touvron, Lavril, Izacard, Martinet, Lachaux, Lacroix, Rozière, Goyal, Hambro, Azhar, Rodriguez, Joulin, Grave, and Lample]{touvron_llama_2023}
Hugo Touvron, Thibaut Lavril, Gautier Izacard, Xavier Martinet, Marie-Anne Lachaux, Timothée Lacroix, Baptiste Rozière, Naman Goyal, Eric Hambro, Faisal Azhar, Aurelien Rodriguez, Armand Joulin, Edouard Grave, and Guillaume Lample.
\newblock {LLaMA}: {Open} and {Efficient} {Foundation} {Language} {Models}, February 2023.
\newblock URL \url{http://arxiv.org/abs/2302.13971}.
\newblock arXiv:2302.13971 [cs].

\bibitem[Moradi et~al.(2021)Moradi, Blagec, Haberl, and Samwald]{moradi_gpt-3_2021}
Milad Moradi, Kathrin Blagec, Florian Haberl, and Matthias Samwald.
\newblock {GPT}-3 {Models} are {Poor} {Few}-{Shot} {Learners} in the {Biomedical} {Domain}, September 2021.
\newblock URL \url{https://arxiv.org/abs/2109.02555v2}.

\bibitem[Tunstall et~al.(2022)Tunstall, Reimers, Jo, Bates, Korat, Wasserblat, and Pereg]{tunstall_efficient_2022}
Lewis Tunstall, Nils Reimers, Unso Eun~Seo Jo, Luke Bates, Daniel Korat, Moshe Wasserblat, and Oren Pereg.
\newblock Efficient {Few}-{Shot} {Learning} {Without} {Prompts}, September 2022.
\newblock URL \url{http://arxiv.org/abs/2209.11055}.
\newblock arXiv:2209.11055 [cs].

\bibitem[Gutiérrez et~al.(2022)Gutiérrez, McNeal, Washington, Chen, Li, Sun, and Su]{gutierrez_thinking_2022}
Bernal~Jiménez Gutiérrez, Nikolas McNeal, Clay Washington, You Chen, Lang Li, Huan Sun, and Yu~Su.
\newblock Thinking about {GPT}-3 {In}-{Context} {Learning} for {Biomedical} {IE}? {Think} {Again}, November 2022.
\newblock URL \url{http://arxiv.org/abs/2203.08410}.
\newblock arXiv:2203.08410 [cs].

\bibitem[Sun et~al.(2023)Sun, Li, Li, Wu, Guo, Zhang, and Wang]{sun_text_2023}
Xiaofei Sun, Xiaoya Li, Jiwei Li, Fei Wu, Shangwei Guo, Tianwei Zhang, and Guoyin Wang.
\newblock Text {Classification} via {Large} {Language} {Models}, May 2023.
\newblock URL \url{http://arxiv.org/abs/2305.08377}.
\newblock arXiv:2305.08377 [cs].

\bibitem[Tang et~al.(2023)Tang, Han, Jiang, and Hu]{tang_does_2023}
Ruixiang Tang, Xiaotian Han, Xiaoqian Jiang, and Xia Hu.
\newblock Does {Synthetic} {Data} {Generation} of {LLMs} {Help} {Clinical} {Text} {Mining}?, April 2023.
\newblock URL \url{http://arxiv.org/abs/2303.04360}.
\newblock arXiv:2303.04360 [cs].

\bibitem[Rohanian et~al.(2023{\natexlab{a}})Rohanian, Nouriborji, and Clifton]{rohanian_exploring_2023}
Omid Rohanian, Mohammadmahdi Nouriborji, and David~A. Clifton.
\newblock Exploring the {Effectiveness} of {Instruction} {Tuning} in {Biomedical} {Language} {Processing}, December 2023{\natexlab{a}}.
\newblock URL \url{http://arxiv.org/abs/2401.00579}.
\newblock arXiv:2401.00579 [cs].

\bibitem[Chen et~al.(2023)Chen, Du, Hu, Keloth, Peng, Raja, Zhang, Lu, and Xu]{chen_large_2023}
Qingyu Chen, Jingcheng Du, Yan Hu, Vipina~Kuttichi Keloth, Xueqing Peng, Kalpana Raja, Rui Zhang, Zhiyong Lu, and Hua Xu.
\newblock Large language models in biomedical natural language processing: benchmarks, baselines, and recommendations, May 2023.
\newblock URL \url{https://arxiv.org/abs/2305.16326v1}.

\bibitem[Hinton et~al.(2015)Hinton, Vinyals, and Dean]{hinton_distilling_2015}
Geoffrey Hinton, Oriol Vinyals, and Jeff Dean.
\newblock Distilling the {Knowledge} in a {Neural} {Network}, March 2015.
\newblock URL \url{http://arxiv.org/abs/1503.02531}.
\newblock arXiv:1503.02531 [cs, stat].

\bibitem[Sanh et~al.(2020)Sanh, Debut, Chaumond, and Wolf]{sanh_distilbert_2020}
Victor Sanh, Lysandre Debut, Julien Chaumond, and Thomas Wolf.
\newblock {DistilBERT}, a distilled version of {BERT}: smaller, faster, cheaper and lighter, February 2020.
\newblock URL \url{http://arxiv.org/abs/1910.01108}.
\newblock arXiv:1910.01108 [cs].

\bibitem[Sun et~al.(2020)Sun, Yu, Song, Liu, Yang, and Zhou]{sun_mobilebert_2020}
Zhiqing Sun, Hongkun Yu, Xiaodan Song, Renjie Liu, Yiming Yang, and Denny Zhou.
\newblock {MobileBERT}: a {Compact} {Task}-{Agnostic} {BERT} for {Resource}-{Limited} {Devices}.
\newblock In \emph{Proceedings of the 58th {Annual} {Meeting} of the {Association} for {Computational} {Linguistics}}, pages 2158--2170, Online, July 2020. Association for Computational Linguistics.
\newblock \doi{10.18653/v1/2020.acl-main.195}.
\newblock URL \url{https://aclanthology.org/2020.acl-main.195}.

\bibitem[Frantar and Alistarh(2023)]{frantar_sparsegpt_2023}
Elias Frantar and Dan Alistarh.
\newblock {SparseGPT}: {Massive} {Language} {Models} {Can} {Be} {Accurately} {Pruned} in {One}-{Shot}, March 2023.
\newblock URL \url{http://arxiv.org/abs/2301.00774}.
\newblock arXiv:2301.00774 [cs].

\bibitem[Luo et~al.(2023)Luo, Yang, Meng, Li, Zhou, and Zhang]{luo_empirical_2023}
Yun Luo, Zhen Yang, Fandong Meng, Yafu Li, Jie Zhou, and Yue Zhang.
\newblock An {Empirical} {Study} of {Catastrophic} {Forgetting} in {Large} {Language} {Models} {During} {Continual} {Fine}-tuning, August 2023.
\newblock URL \url{http://arxiv.org/abs/2308.08747}.
\newblock arXiv:2308.08747 [cs].

\bibitem[Devlin et~al.(2019)Devlin, Chang, Lee, and Toutanova]{devlin_bert_2019}
Jacob Devlin, Ming-Wei Chang, Kenton Lee, and Kristina Toutanova.
\newblock {BERT}: {Pre}-training of {Deep} {Bidirectional} {Transformers} for {Language} {Understanding}, May 2019.
\newblock URL \url{http://arxiv.org/abs/1810.04805}.
\newblock arXiv:1810.04805 [cs].

\bibitem[Dettmers et~al.(2023)Dettmers, Pagnoni, Holtzman, and Zettlemoyer]{dettmers_qlora_2023}
Tim Dettmers, Artidoro Pagnoni, Ari Holtzman, and Luke Zettlemoyer.
\newblock {QLoRA}: {Efficient} {Finetuning} of {Quantized} {LLMs}, May 2023.
\newblock URL \url{https://arxiv.org/abs/2305.14314v1}.

\bibitem[Lester et~al.(2021)Lester, Al-Rfou, and Constant]{lester_power_2021}
Brian Lester, Rami Al-Rfou, and Noah Constant.
\newblock The {Power} of {Scale} for {Parameter}-{Efficient} {Prompt} {Tuning}.
\newblock April 2021.
\newblock URL \url{http://arxiv.org/abs/2104.08691}.

\bibitem[Li and Liang(2021)]{li_prefix-tuning_2021}
Xiang~Lisa Li and Percy Liang.
\newblock Prefix-{Tuning}: {Optimizing} {Continuous} {Prompts} for {Generation}, January 2021.
\newblock URL \url{http://arxiv.org/abs/2101.00190}.
\newblock arXiv:2101.00190 [cs].

\bibitem[Hu et~al.(2021)Hu, Shen, Wallis, Allen-Zhu, Li, Wang, Wang, and Chen]{hu_lora_2021}
Edward~J. Hu, Yelong Shen, Phillip Wallis, Zeyuan Allen-Zhu, Yuanzhi Li, Shean Wang, Lu~Wang, and Weizhu Chen.
\newblock {LoRA}: {Low}-{Rank} {Adaptation} of {Large} {Language} {Models}, October 2021.
\newblock URL \url{http://arxiv.org/abs/2106.09685}.
\newblock arXiv:2106.09685 [cs].

\bibitem[Liu et~al.(2022)Liu, Tam, Muqeeth, Mohta, Huang, Bansal, and Raffel]{liu_few-shot_2022}
Haokun Liu, Derek Tam, Mohammed Muqeeth, Jay Mohta, Tenghao Huang, Mohit Bansal, and Colin Raffel.
\newblock Few-{Shot} {Parameter}-{Efficient} {Fine}-{Tuning} is {Better} and {Cheaper} than {In}-{Context} {Learning}, August 2022.
\newblock URL \url{http://arxiv.org/abs/2205.05638}.
\newblock arXiv:2205.05638 [cs].

\bibitem[Alsentzer et~al.(2019)Alsentzer, Murphy, Boag, Weng, Jin, Naumann, and McDermott]{alsentzer_publicly_2019}
Emily Alsentzer, John~R. Murphy, Willie Boag, Wei-Hung Weng, Di~Jin, Tristan Naumann, and Matthew B.~A. McDermott.
\newblock Publicly {Available} {Clinical} {BERT} {Embeddings}.
\newblock April 2019.
\newblock URL \url{http://arxiv.org/abs/1904.03323}.

\bibitem[Lehman et~al.(2023)Lehman, Hernandez, Mahajan, Wulff, Smith, Ziegler, Nadler, Szolovits, Johnson, and Alsentzer]{lehman_we_2023}
Eric Lehman, Evan Hernandez, Diwakar Mahajan, Jonas Wulff, Micah~J. Smith, Zachary Ziegler, Daniel Nadler, Peter Szolovits, Alistair Johnson, and Emily Alsentzer.
\newblock Do {We} {Still} {Need} {Clinical} {Language} {Models}?, February 2023.
\newblock URL \url{http://arxiv.org/abs/2302.08091}.
\newblock arXiv:2302.08091 [cs].

\bibitem[Yu et~al.(2023)Yu, Yang, Pelrine, Godbout, and Rabbany]{yu_open_2023}
Hao Yu, Zachary Yang, Kellin Pelrine, Jean~Francois Godbout, and Reihaneh Rabbany.
\newblock Open, {Closed}, or {Small} {Language} {Models} for {Text} {Classification}?, August 2023.
\newblock URL \url{https://arxiv.org/abs/2308.10092v1}.

\bibitem[Rohanian et~al.(2023{\natexlab{b}})Rohanian, Nouriborji, Jauncey, Kouchaki, Group, Clifton, Merson, and Clifton]{rohanian_lightweight_2023}
Omid Rohanian, Mohammadmahdi Nouriborji, Hannah Jauncey, Samaneh Kouchaki, ISARIC Clinical~Characterisation Group, Lei Clifton, Laura Merson, and David~A. Clifton.
\newblock Lightweight {Transformers} for {Clinical} {Natural} {Language} {Processing}, February 2023{\natexlab{b}}.
\newblock URL \url{http://arxiv.org/abs/2302.04725}.
\newblock arXiv:2302.04725 [cs].

\bibitem[Ding et~al.(2023)Ding, Qin, Yang, Wei, Yang, Su, Hu, Chen, Chan, Chen, Yi, Zhao, Wang, Liu, Zheng, Chen, Liu, Tang, Li, and Sun]{ding_parameter-efficient_2023}
Ning Ding, Yujia Qin, Guang Yang, Fuchao Wei, Zonghan Yang, Yusheng Su, Shengding Hu, Yulin Chen, Chi-Min Chan, Weize Chen, Jing Yi, Weilin Zhao, Xiaozhi Wang, Zhiyuan Liu, Hai-Tao Zheng, Jianfei Chen, Yang Liu, Jie Tang, Juanzi Li, and Maosong Sun.
\newblock Parameter-efficient fine-tuning of large-scale pre-trained language models.
\newblock \emph{Nature Machine Intelligence}, 5\penalty0 (3):\penalty0 220--235, March 2023.
\newblock ISSN 2522-5839.
\newblock \doi{10.1038/s42256-023-00626-4}.
\newblock URL \url{https://www.nature.com/articles/s42256-023-00626-4}.
\newblock Number: 3 Publisher: Nature Publishing Group.

\bibitem[Gema et~al.(2023)Gema, Daines, Minervini, and Alex]{gema_parameter-efficient_2023}
Aryo~Pradipta Gema, Luke Daines, Pasquale Minervini, and Beatrice Alex.
\newblock Parameter-{Efficient} {Fine}-{Tuning} of {LLaMA} for the {Clinical} {Domain}, July 2023.
\newblock URL \url{http://arxiv.org/abs/2307.03042}.
\newblock arXiv:2307.03042 [cs].

\bibitem[Jiao et~al.(2020)Jiao, Yin, Shang, Jiang, Chen, Li, Wang, and Liu]{jiao_tinybert_2020}
Xiaoqi Jiao, Yichun Yin, Lifeng Shang, Xin Jiang, Xiao Chen, Linlin Li, Fang Wang, and Qun Liu.
\newblock {TinyBERT}: {Distilling} {BERT} for {Natural} {Language} {Understanding}, October 2020.
\newblock URL \url{http://arxiv.org/abs/1909.10351}.
\newblock arXiv:1909.10351 [cs].

\bibitem[Rohanian et~al.(2023{\natexlab{c}})Rohanian, Nouriborji, Kouchaki, and Clifton]{rohanian_effectiveness_2023}
Omid Rohanian, Mohammadmahdi Nouriborji, Samaneh Kouchaki, and David~A Clifton.
\newblock On the effectiveness of compact biomedical transformers.
\newblock \emph{Bioinformatics}, 39\penalty0 (3):\penalty0 btad103, March 2023{\natexlab{c}}.
\newblock ISSN 1367-4811.
\newblock \doi{10.1093/bioinformatics/btad103}.
\newblock URL \url{https://doi.org/10.1093/bioinformatics/btad103}.

\bibitem[Taylor et~al.(2023)Taylor, Zhang, Joyce, Gao, Kormilitzin, and Nevado-Holgado]{taylor_clinical_2023}
Niall Taylor, Yi~Zhang, Dan~W. Joyce, Ziming Gao, Andrey Kormilitzin, and Alejo Nevado-Holgado.
\newblock Clinical {Prompt} {Learning} {With} {Frozen} {Language} {Models}.
\newblock \emph{IEEE Transactions on Neural Networks and Learning Systems}, pages 1--11, 2023.
\newblock ISSN 2162-2388.
\newblock \doi{10.1109/TNNLS.2023.3294633}.
\newblock URL \url{https://ieeexplore.ieee.org/document/10215061}.

\bibitem[Meehan et~al.(2022)Meehan, Lewis, Fazel, Fusar-Poli, Steyerberg, Stahl, and Danese]{meehan_clinical_2022}
Alan~J. Meehan, Stephanie~J. Lewis, Seena Fazel, Paolo Fusar-Poli, Ewout~W. Steyerberg, Daniel Stahl, and Andrea Danese.
\newblock Clinical prediction models in psychiatry: a systematic review of two decades of progress and challenges.
\newblock \emph{Molecular Psychiatry}, 27\penalty0 (6):\penalty0 2700--2708, June 2022.
\newblock ISSN 1476-5578.
\newblock \doi{10.1038/s41380-022-01528-4}.
\newblock URL \url{https://www.nature.com/articles/s41380-022-01528-4}.
\newblock Number: 6 Publisher: Nature Publishing Group.

\bibitem[Lee et~al.(2020)Lee, Yoon, Kim, Kim, Kim, So, and Kang]{lee_biobert_2020}
Jinhyuk Lee, Wonjin Yoon, Sungdong Kim, Donghyeon Kim, Sunkyu Kim, Chan~Ho So, and Jaewoo Kang.
\newblock {BioBERT}: {A} pre-trained biomedical language representation model for biomedical text mining.
\newblock \emph{Bioinformatics}, 36\penalty0 (4):\penalty0 1234--1240, February 2020.
\newblock \doi{10.1093/bioinformatics/btz682}.
\newblock Publisher: Oxford University Press.

\bibitem[Johnson et~al.(2016)Johnson, Pollard, Shen, Lehman, Feng, Ghassemi, Moody, Szolovits, Anthony~Celi, and Mark]{johnson_mimic-iii_2016}
Alistair~E.W. Johnson, Tom~J. Pollard, Lu~Shen, Li~Wei~H. Lehman, Mengling Feng, Mohammad Ghassemi, Benjamin Moody, Peter Szolovits, Leo Anthony~Celi, and Roger~G. Mark.
\newblock {MIMIC}-{III}, a freely accessible critical care database.
\newblock \emph{Scientific Data}, 3, May 2016.
\newblock \doi{10.1038/sdata.2016.35}.
\newblock Publisher: Nature Publishing Groups.

\bibitem[Van~Aken et~al.(2021)Van~Aken, Papaioannou, Mayrdorfer, Budde, Gers, and Loeser]{van_aken_clinical_2021}
Betty Van~Aken, Jens-Michalis Papaioannou, Manuel Mayrdorfer, Klemens Budde, Felix Gers, and Alexander Loeser.
\newblock Clinical {Outcome} {Prediction} from {Admission} {Notes} using {Self}-{Supervised} {Knowledge} {Integration}.
\newblock In \emph{Proceedings of the 16th {Conference} of the {European} {Chapter} of the {Association} for {Computational} {Linguistics}: {Main} {Volume}}, pages 881--893, Online, 2021. Association for Computational Linguistics.
\newblock \doi{10.18653/v1/2021.eacl-main.75}.
\newblock URL \url{https://aclanthology.org/2021.eacl-main.75}.

\bibitem[Uzuner et~al.(2011)Uzuner, South, Shen, and DuVall]{uzuner_2010_2011}
Özlem Uzuner, Brett~R South, Shuying Shen, and Scott~L DuVall.
\newblock 2010 i2b2/{VA} challenge on concepts, assertions, and relations in clinical text.
\newblock \emph{Journal of the American Medical Informatics Association : JAMIA}, 18\penalty0 (5):\penalty0 552--556, 2011.
\newblock ISSN 1067-5027.
\newblock \doi{10.1136/amiajnl-2011-000203}.
\newblock URL \url{https://www.ncbi.nlm.nih.gov/pmc/articles/PMC3168320/}.

\bibitem[Sun et~al.(2013)Sun, Rumshisky, and Uzuner]{sun_evaluating_2013}
Weiyi Sun, Anna Rumshisky, and Ozlem Uzuner.
\newblock Evaluating temporal relations in clinical text: 2012 i2b2 {Challenge}.
\newblock \emph{Journal of the American Medical Informatics Association : JAMIA}, 20\penalty0 (5):\penalty0 806--813, September 2013.
\newblock ISSN 1067-5027.
\newblock \doi{10.1136/amiajnl-2013-001628}.
\newblock URL \url{https://www.ncbi.nlm.nih.gov/pmc/articles/PMC3756273/}.

\bibitem[Stubbs et~al.(2015)Stubbs, Kotfila, and Uzuner]{stubbs_automated_2015}
Amber Stubbs, Christopher Kotfila, and Özlem Uzuner.
\newblock Automated systems for the de-identification of longitudinal clinical narratives: {Overview} of 2014 i2b2/{UTHealth} shared task {Track} 1.
\newblock \emph{Journal of Biomedical Informatics}, 58:\penalty0 S11--S19, December 2015.
\newblock ISSN 1532-0464.
\newblock \doi{10.1016/j.jbi.2015.06.007}.
\newblock URL \url{https://www.sciencedirect.com/science/article/pii/S1532046415001173}.

\bibitem[Akiba et~al.(2019)Akiba, Sano, Yanase, Ohta, and Koyama]{akiba_optuna_2019}
Takuya Akiba, Shotaro Sano, Toshihiko Yanase, Takeru Ohta, and Masanori Koyama.
\newblock Optuna: {A} {Next}-generation {Hyperparameter} {Optimization} {Framework}, July 2019.
\newblock URL \url{http://arxiv.org/abs/1907.10902}.
\newblock arXiv:1907.10902 [cs, stat].

\bibitem[Mangrulkar et~al.(2022)Mangrulkar, Gugger, Debut, Belkada, Paul, and Bossan]{peft}
Sourab Mangrulkar, Sylvain Gugger, Lysandre Debut, Younes Belkada, Sayak Paul, and Benjamin Bossan.
\newblock Peft: State-of-the-art parameter-efficient fine-tuning methods.
\newblock \url{https://github.com/huggingface/peft}, 2022.

\bibitem[Wolf et~al.(2020)Wolf, Debut, Sanh, Chaumond, Delangue, Moi, Cistac, Rault, Louf, Funtowicz, Davison, Shleifer, von Platen, Ma, Jernite, Plu, Xu, Le~Scao, Gugger, Drame, Lhoest, and Rush]{wolf-etal-2020-transformers}
Thomas Wolf, Lysandre Debut, Victor Sanh, Julien Chaumond, Clement Delangue, Anthony Moi, Pierric Cistac, Tim Rault, Remi Louf, Morgan Funtowicz, Joe Davison, Sam Shleifer, Patrick von Platen, Clara Ma, Yacine Jernite, Julien Plu, Canwen Xu, Teven Le~Scao, Sylvain Gugger, Mariama Drame, Quentin Lhoest, and Alexander Rush.
\newblock Transformers: State-of-the-art natural language processing.
\newblock In \emph{Proceedings of the 2020 Conference on Empirical Methods in Natural Language Processing: System Demonstrations}, pages 38--45, Online, October 2020. Association for Computational Linguistics.
\newblock \doi{10.18653/v1/2020.emnlp-demos.6}.
\newblock URL \url{https://aclanthology.org/2020.emnlp-demos.6}.

\end{thebibliography}

\appendix
\section{Dataset details}
\label{appendix:dataset-details}
\subsection{MIMIC-III}
Mimic-III is a large, freely-available database comprising deidentified health data associated with over 40,000 patients who stayed in critical care units of the Beth Israel Deaconess Medical Center between 2001 and 2012 \cite{johnson_mimic-iii_2016}.The data includes demographics, vital signs, laboratory tests, medications, and more collected from a variety of hospital systems. It encompasses over 2 million notes including discharge summaries, radiology reports, and more.

\subsection{i2b2}
Originally released on the \href{https://www.i2b2.org}{i2b2 website}, but is now hosted via the Department of \href{https://portal.dbmi.hms.harvard.edu/}{BioMedical Informatics (DBMI) data portal}. The dataset is now referred to as the National NLP Clinical Challenges research datasets (n2c2), and is based on fully deidentified notes from the Research Patient Data Registry at Partners Healthcare System in Boston. 

\section{LoRA Rank Analysis}
\label{appendix:lora-rank}
\begin{figure}
    \centering
    \includegraphics[scale=0.5]{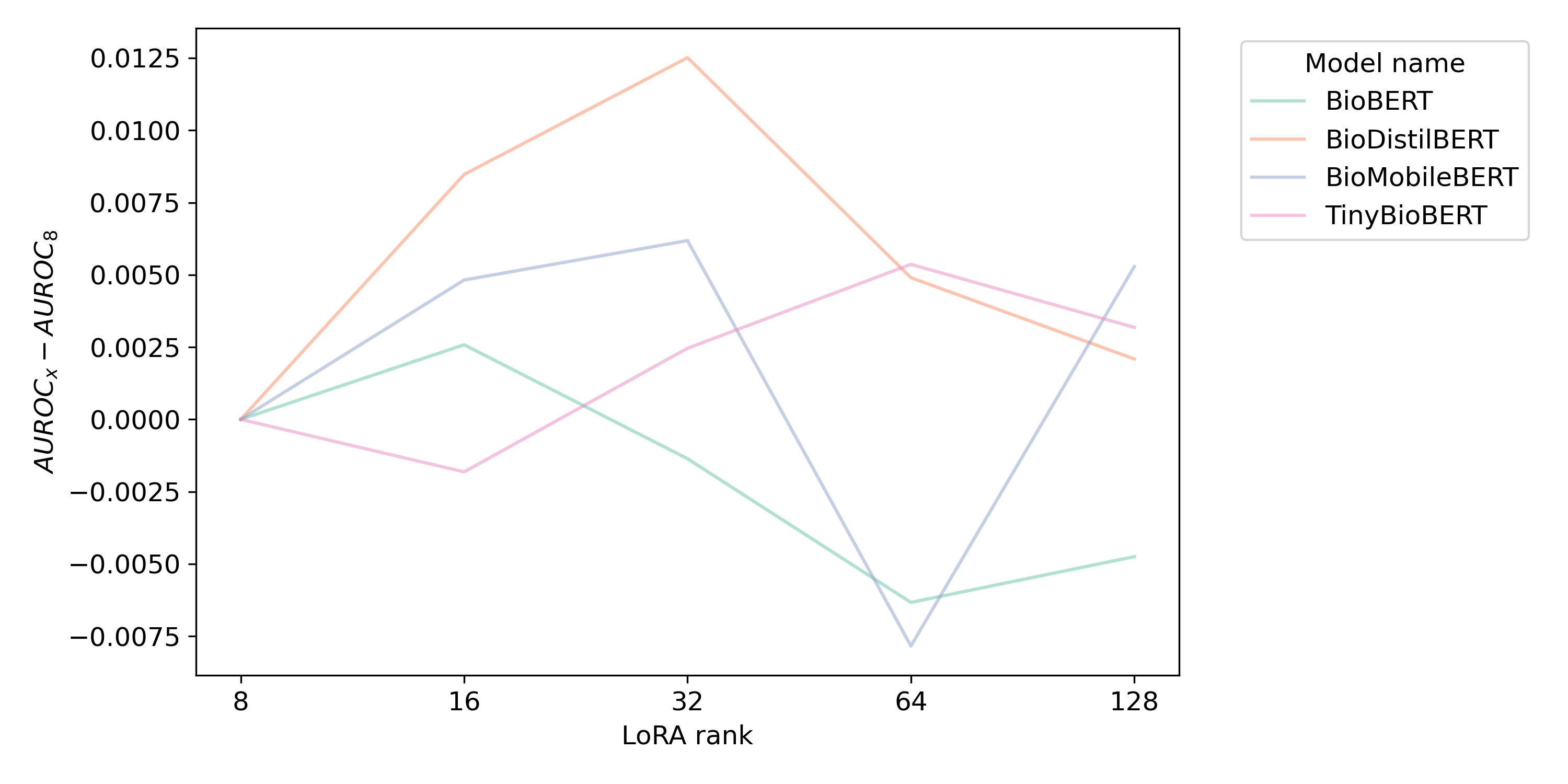}
    \caption{Differential effect of LoRA rank on performance of a model. The y-axis represents the difference in AUROC between the rank on the x-axis and rank=8.}
    \label{appendix:LoRA-rank-vs-performance}
\end{figure}

We provide a comparison of different LoRA ranks on task performance across each model in Figure \ref{appendix:LoRA-rank-vs-performance}.

\section{Hyperparameters and hardware for downstream tasks}

For the core experiments we utilised the HuggingFace\cite{wolf-etal-2020-transformers} and Parameter Efficient Finetuning (PEFT)\cite{peft} libraries. For consistency and equal footing between model types, all experiments utilised a single NVIDIA RTX 3090 graphics card with 24GB of VRAM. Due to this, however, the experiments utilising Llama-2-7b, even with LoRA, required a reduction in the precision of the model weights from fp32 to bfloat16. 

\label{appendix:training-hyperparams}

\begin{table}[htp]
    \centering
    \footnotesize
    \begin{tabular}{llc}
        \toprule
        PEFT & Hyperparameter & Value \\
        \midrule
        \multirow{3}{*}{LoRA} & r & 8 \\
                              & alpha & 8 \\
                              & dropout & 0.1 \\
                              & learning rate & $3e-4$ \\
                              & target modules & [key, value] \\
                              & layers & all \\
        \midrule
        \multirow{3}{*}{$IA^3$} & dropout & 0.1 \\
                                & learning rate & $3e-4$\\
                                & target modules & [key, value, feed-forward] \\
                                & layers & all \\

        \bottomrule
    \end{tabular}
    \caption{The default hyperparameters for LoRA and $IA^3$ used in all experiments prior to the hyperparameter optimisation. For full fine-tuning the same learning rate ($3e-4$) and dropout (0.1) was used.}
    \label{appendix:tab-training-hyperparams}
\end{table}

\begin{table}[htp]

\centering
\begin{subtable}{\textwidth}
  \centering
  \footnotesize
  \begin{tabular}{llrrrr}
    \toprule  
        \multicolumn{1}{c}{\bf Model name} & \bf PEFT & \multicolumn{1}{c}{\bf ICD9-Triage} & \multicolumn{1}{c}{\bf i2b2-2010-RE} & \multicolumn{1}{c}{\bf MIMIC-LoS} & \multicolumn{1}{c}{\bf Mimic-MP} \\
        \midrule
        BERTbase & Full & 0.991 & 0.975 & 0.702 & 0.799 \\
        BERTbase & LORA & 0.983 & 0.980 & 0.679 & 0.811 \\
        \midrule
        BioBERT & Full & 0.991 & 0.982 & 0.711 & 0.812 \\  
        BioBERT & LORA & 0.991 & 0.985 & 0.697 & 0.828 \\
        \midrule
        BioClinicalBERT & Full & 0.993 & 0.978 & 0.697 & 0.793 \\
        BioClinicalBERT & LORA & 0.990 & 0.981 & 0.701 & 0.822 \\  
        \midrule
        BioDistilBERT & Full & 0.992 & 0.979 & 0.697 & 0.803 \\
        BioDistilBERT & LORA & 0.993 & 0.988 & 0.704 & 0.822 \\
        \midrule
        BioMobileBERT & Full & 0.992 & 0.980 & 0.697 & 0.809 \\ 
        BioMobileBERT & LORA & 0.987 & 0.982 & 0.670 & 0.792 \\  
        \midrule
        ClinicalDistilBERT & Full & 0.994 & 0.980 & 0.697 & 0.822 \\
        ClinicalDistilBERT & LORA & 0.995 & 0.989 & 0.710 & 0.836 \\
        \midrule  
        ClinicalMobileBERT & Full & 0.995 & 0.983 & 0.720 & 0.826 \\  
        ClinicalMobileBERT & LORA & 0.994 & 0.982 & 0.690 & 0.824 \\
        \bottomrule
  \end{tabular}
  \caption{Sequence classification task results}
  \label{tab:all_domain_cls_results}
\end{subtable}

\vspace{0.5cm} 

\begin{subtable}{\textwidth}
  \centering
  \footnotesize
 \begin{tabular}{llrrr}
   \toprule
        \multicolumn{1}{c}{\bf Model name} & \bf PEFT & \multicolumn{1}{c}{\bf i2b2-2010-NER} & \multicolumn{1}{c}{\bf i2b2-2012-NER} & \multicolumn{1}{c}{\bf i2b2-2014-NER} \\
    \midrule
    BERTbase & Full & 0.806 & 0.792 & 0.974 \\  
    BERTbase & LORA & 0.673 & 0.697 & 0.951 \\
    \midrule
    BioBERT & Full & 0.822 & 0.823 & 0.969 \\
    BioBERT & LORA & 0.713 & 0.757 & 0.935 \\  
    \midrule 
    BioClinicalBERT & Full & 0.846 & 0.820 & 0.960 \\
    BioClinicalBERT & LORA & 0.704 & 0.746 & 0.920 \\
    \midrule
    BioDistilBERT & Full & 0.809 & 0.794 & 0.965 \\ 
    BioDistilBERT & LORA & 0.704 & 0.726 & 0.939 \\  
    \midrule
    BioMobileBERT & Full & 0.794 & 0.774 & 0.966 \\
    BioMobileBERT & LORA & 0.649 & 0.654 & 0.938 \\  
    \midrule
    ClinicalDistilBERT & Full & 0.816 & 0.817 & 0.961 \\ 
    ClinicalDistilBERT & LORA & 0.671 & 0.740 & 0.920 \\ 
    \bottomrule
 \end{tabular}
 \caption{NER task results}
 \label{tab:all_domain_ner_results}
\end{subtable}
\caption{PEFT results for sequence classification and NER tasks dependent on domain pre-training received.} 
\label{appendix:all-domain-results}
\end{table}
\end{document}